\documentclass[11pt]{article}

\usepackage[final]{acl}

\usepackage{times}
\usepackage{latexsym}
\usepackage[most]{tcolorbox} 
\usepackage{enumitem}

\usepackage[T1]{fontenc}

\usepackage[utf8]{inputenc}

\usepackage{microtype}

\usepackage{inconsolata}

\usepackage{graphicx}

\usepackage{booktabs}
\usepackage{tabularx}
\usepackage{float}
\usepackage{hyperref}
\usepackage{lipsum}
\usepackage{multirow}
\usepackage{makecell}
\usepackage{longtable}
\usepackage[table]{xcolor}

\usepackage{pgfplots}
\usepgfplotslibrary{groupplots} 
\usetikzlibrary{shadows, calc} 
\pgfplotsset{compat=1.18}
\usepackage{float}
\definecolor{colorP}{HTML}{66C2A5}
\definecolor{colorPR}{HTML}{FC8D62}
\definecolor{colorPQ}{HTML}{8DA0CB}
\definecolor{cRed}{HTML}{B30021}
\definecolor{cOrange}{HTML}{FF8C42}
\definecolor{cYellow}{HTML}{FFFFB3}
\definecolor{cLightGreen}{HTML}{75D169}
\definecolor{cDarkGreen}{HTML}{006B38}

\title{Text Generation in Low-Resource Languages\\ Using Human-Curated Vocabularies}

\author{
\textbf{Michael Christlambert Sinanta\textsuperscript{1*} \quad
Muhammad Dehan Al Kautsar\textsuperscript{2*}} \\
\textbf{Amanda Nurul Izzah\textsuperscript{1} \quad
Thanh-Nhi Nguyen\textsuperscript{1} \quad
Phan Phuong Mai Chau\textsuperscript{2}} \\
\textbf{Zheng Wei Lim\textsuperscript{3} \quad
Fajri Koto\textsuperscript{2}}\\[4pt]
\textsuperscript{1}Independent Researcher\\
\textsuperscript{2}Mohamed bin Zayed University of Artificial Intelligence\\
\textsuperscript{3}University of Amsterdam\\[2pt]
\texttt{michaelchristlambert@gmail.com, muhammad.dehan@mbzuai.ac.ae}
}

\begin{document}
\maketitle
{\renewcommand{\thefootnote}{}\footnotetext{\textsuperscript{*}Corresponding authors.}}
\begin{abstract}

Synthetic text generation offers a promising approach to alleviating data scarcity in low-resource languages. However, its effectiveness is inherently limited for languages that are scarcely represented, or entirely absent, in an LLM’s pretraining data. At the same time, many such languages have human-curated vocabularies that provide rich lexical and semantic information, often accompanied by grammatical information such as part-of-speech labels. In this work, we ask whether LLMs can leverage these resources to generate coherent text in extremely low-resource and unseen languages. We introduce a framework that combines human-curated vocabularies with a small number of human-written exemplars that provide syntactic signals, while sampling candidate words according to a Zipfian distribution. Across languages with varying levels of resource availability, we evaluate the generated text using lexical diversity, off-target ratio, and human judgments of fluency. Our results show that reasoning-enabled LLMs can generate fluent text even in extremely low-resource languages, with particularly strong results at the sentence level and promising performance at the paragraph level. Our findings highlight a practical pathway for bootstrapping synthetic data in low-resource languages, expanding the amount of usable text even when existing digital resources are extremely limited.\footnote{Our source code is available at \url{https://github.com/michaelsinanta/vocab2text}.}
\end{abstract}

\section{Introduction}

\begin{figure}[ht]
    \centering
    \includegraphics[width=\columnwidth]{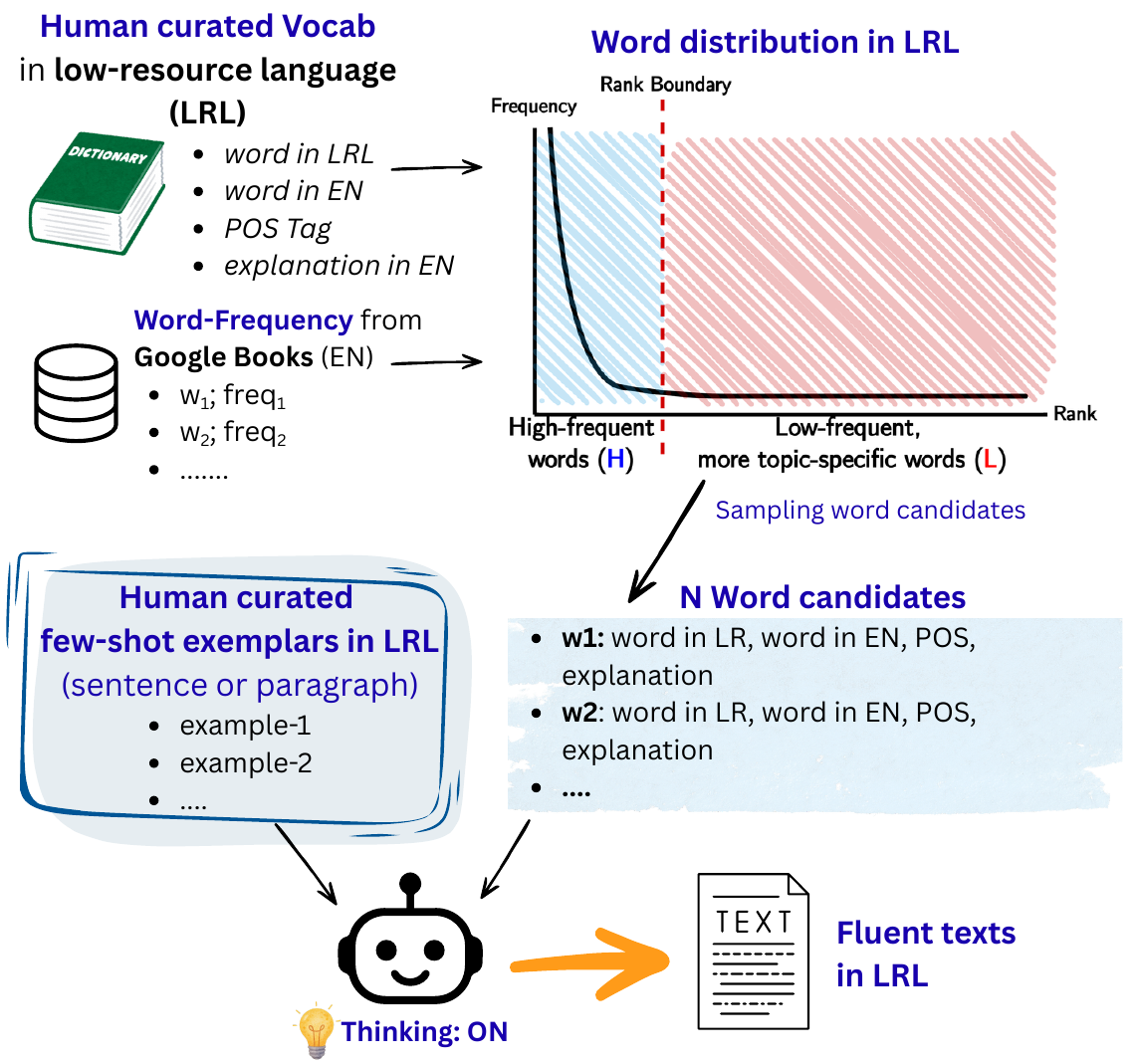}
    \caption{Overview of our vocabulary-guided synthetic text generation pipeline. Human-curated vocabulary is sampled using an open-source word-frequency list, while we then sample the \textcolor{blue}{high-} and \textcolor{red}{low-}frequency words using our method described in Section~\ref{sec:methodology}. We then combine it with few-shot human-written exemplars to construct the generation prompt.}
    \label{fig:methodology}
\end{figure}

More than 7,000 languages are spoken worldwide, yet only a small fraction are represented in the textual resources commonly used to train language models. For example, Wikipedia and Common Crawl contain textual data for only about 4\% of the world's languages, while human-curated lexicons and word lists cover approximately 70\%~\citep{wang-etal-2022-expanding}. This broader coverage reflects longstanding language documentation efforts, through which linguists and language communities have compiled human-curated vocabularies that record words, their meanings, and other linguistic information. Their broad coverage creates an opportunity to leverage these existing resources to expand text resources for low-resource languages.

One promising way to realize this opportunity is through synthetic text generation. LLMs are increasingly used to generate synthetic data for pretraining and post-training~\citep{wang-etal-2023-self-instruct, doshi-etal-2024-pretraining, anikina-etal-2025-rigorous, stenlund-etal-2026-synthetic}. However, such generation largely relies on the linguistic knowledge already encoded in LLMs, limiting its effectiveness for languages that are poorly represented in their pretraining data~\citep{guo-etal-2024-curious, li-etal-2025-context-learning}. Human-curated vocabularies offer a complementary source of linguistic knowledge, yet their potential for LLM-based synthetic text generation remains largely unexplored~\citep{yong-etal-2024-lexc}. This idea has a natural parallel in human language learning: when learning an unfamiliar language, people consult dictionaries to acquire new words and use examples to learn how those words appear in context. Inspired by this process, we ask: \textit{Can LLMs leverage human-curated vocabularies to generate coherent and lexically diverse text in low-resource languages, including extremely underrepresented languages?}

Our contributions are threefold: \begin{enumerate}[itemsep=1pt, topsep=2pt, parsep=0pt, partopsep=0pt] \item We introduce a training-free framework that leverages LLM reasoning to generate text in low-resource languages from human-curated vocabularies and a small number of human-written exemplars. Our framework adapts from Zipfian sampling to expose the model to a broad yet natural distribution of vocabulary during generation. \item Across six languages spanning different levels of resource availability, our framework consistently improves lexical diversity and reduces off-target generation over practical baselines. Human evaluation further shows improved fluency, with particularly strong results for the extremely low-resource Acehnese and Hitu. \item Through extensive analyses, we examine when and why vocabulary-guided generation works, studying the impact of vocabulary quality, reasoning versus non-reasoning generation, and increasing generation volume. Our results highlight vocabulary quality as a key consideration and show whether gains in lexical diversity persist as synthetic data are generated at larger scale. \end{enumerate}

\section{Related Works}

\paragraph{Synthetic Text Generation with LLMs.} Synthetic data has long been used to supplement scarce human-written data in NLP~\citep{feng-etal-2021-survey}. Early augmentation methods typically transform existing examples through operations such as synonym replacement, random insertion and deletion, and word swapping~\citep{wei-zou-2019-eda, karimi-etal-2021-aeda-easier}. 
The emergence of LLMs has enabled a shift toward generating new synthetic text at scale, with applications ranging from augmenting pretraining corpora~\citep{doshi-etal-2024-pretraining,zhu-etal-2026-synthetic} to constructing post-training resources such as instruction-tuning data~\citep{wang-etal-2023-self-instruct}. Despite their strong generation capabilities, LLMs tend to produce text with lower lexical diversity than human-written data~\citep{guo-etal-2024-curious,kambhatla-etal-2025-measuring}. 
This limitation is particularly relevant to our goal of expanding the lexical coverage of languages with limited textual resources. Therefore, we treat lexical diversity, alongside fluency, as a central dimension of synthetic text quality.

\paragraph{Synthetic Data for Low-Resource Languages.} Synthetic text generation is particularly challenging for low-resource languages, where LLMs have limited linguistic knowledge. Generation quality tends to deteriorate as language resources decrease~\citep{bafna-etal-2025-translation}, and models may drift into higher-resource languages when generating underrepresented ones~\citep{marchisio-etal-2024-understanding}. Nevertheless, LLM-generated data have been explored for low-resource applications ranging from question answering to story generation~\citep{namboori2023gemquad, cegin-etal-2025-use, pranida-etal-2025-culturally}, with target-language demonstrations shown to improve their utility~\citep{anikina-etal-2025-rigorous}. These approaches, however, largely rely on linguistic knowledge already encoded in the LLM. Our work instead asks whether LLMs can generate fluent and lexically diverse text with minimal off-target generation in extremely underrepresented languages, including languages not seen during LLM pretraining, 
where we leverage LLM reasoning to combine with external lexical knowledge to provide additional linguistic information at inference time.

\paragraph{Human-Curated Vocabularies in NLP.} Human-curated lexical resources, including dictionaries, bilingual lexicons, and word lists, cover substantially more languages than commonly available text corpora~\citep{wang-etal-2022-expanding}. Prior work has leveraged these resources for lexical augmentation, word-for-word data synthesis, and machine translation~\citep{feng-etal-2021-survey, wang-etal-2022-expanding, peng2020dictionarybaseddataaugmentationcrossdomain}. Recent studies further show that providing dictionaries or grammatical resources in context can help LLMs translate languages scarcely represented in pretraining~\citep{tanzer-etal-2024-benchmark, zhang-etal-2024-teaching}. These approaches, however, primarily focus on translation. 
Closest to our work, \citet{yong-etal-2024-lexc} condition LLMs on bilingual lexicons to generate task-specific data, which are subsequently translated word by word into the target language. In contrast, we study open-ended generation directly in the target language. 
We use human-curated vocabularies as lexical knowledge and a few human-written exemplars as syntactic guidance, requiring the LLM to compose new text rather than translate or modify existing sentences.

\section{Methodology} 
\label{sec:methodology}

We illustrate our training-free, vocabulary-guided synthetic text generation pipeline in Figure~\ref{fig:methodology}. The pipeline consists of four stages: (1) collecting human-curated dictionaries and word-frequency information for each target language, (2) collecting a small set of human-written exemplars in sentences and paragraphs, (3) sampling candidate words according to their frequency distribution, and (4) constructing the generation prompt from the sampled words and exemplars.

\paragraph{Stage 1: Collecting Dictionaries and Estimating Word Frequencies.}
\label{sec:subsec_dict_word_freq_collection}
We first collect a human-curated dictionary $\mathcal{D}$ for each target language and standardize its entries into a common format. Each entry $w \in \mathcal{D}$ consists of a target-language word, its English equivalent $\mathrm{en}(w)$, an English definition, a part-of-speech (POS) tag, and, when available, an example sentence illustrating its usage. 
Because the available resources differ substantially across languages and formats, we use language-specific procedures to construct the dictionaries. 
Appendix~\ref{sec:appendix_dictionary_resources} describes the data sources and processing procedures and reports the number of entries collected for each language (Table~\ref{tab:dictionary_size}).

We next estimate how frequently each dictionary word is likely to occur in natural text. Direct frequency estimates are unavailable or unreliable for many of our target languages, particularly the extremely low-resource ones. We therefore derive approximate frequencies from English using the Dictionary Word Frequency dataset,\footnote{A preprocessed~\href{https://www.kaggle.com/datasets/wheelercode/dictionary-word-frequency}{dataset} based on~\href{https://storage.googleapis.com/books/ngrams/books/datasetsv3.html}{Google Books Ngram}.} which provides unigram counts from Google Books. Each entry $w$ is assigned the frequency $f(w)$ of its English equivalent $\mathrm{en}(w)$. Entries whose English equivalent has no frequency are excluded from frequency-based sampling, and the remaining entries form the vocabulary $\mathcal{V} \subseteq \mathcal{D}$. We rank the entries of $\mathcal{V}$ from highest to lowest $f(w)$ and denote the rank of entry $w$ by $r(w)$, so that $r(w) = 1$ for the most frequent entry. The resulting distribution is heavy-tailed, with a small number of highly frequent words and a long tail of less frequent words, consistent with Zipf's law. We use this ranking for word-candidate sampling in Stage~3.

\paragraph{Stage 2: Collecting Few-Shot Exemplars.}
In parallel, we collect a small set of human-written exemplars to provide syntactic and contextual guidance for generation~\citep{anikina-etal-2025-rigorous}. For each language (Section~\ref{sec:exp_setup}), we recruit one native-speaker annotator. Recruiting native speakers is particularly challenging for the extremely low-resource languages in our study. To ensure the quality of the exemplars, we apply strict eligibility criteria: annotators must have at least a senior high school education, must have lived for at least 10 years in a community where the target language is spoken, and must have at least one parent who speaks the language. Each annotator writes 40 sentences and 5 paragraphs. The sentences cover four target lengths (5, 10, 15, and 20 words), with 10 sentences per length. Each paragraph contains at least three sentences and at least 25 words. We denote the resulting pool of exemplars by $\mathcal{E}$. Each exemplar $e \in \mathcal{E}$ consists of a target-language text, its English translation, and the set of dictionary entries $\mathcal{D}(e) \subseteq \mathcal{D}$ used to compose it, following the schema of Stage~1. These exemplars show how dictionary entries are composed naturally in context, and we use them as few-shot guidance during generation. Appendix~\ref{sec:appendix_few_shot_exemplar_creation_procedure} describes the annotation procedure in detail.

\paragraph{Stage 3: Sampling Word Candidates.}
\label{sec:subsec_word_candidates_sampling}
To generate each synthetic sentence or paragraph, we provide the LLM with two inputs: $s$ few-shot exemplars and $k$ word candidates. The exemplars provide syntactic and contextual guidance, while the word candidates encourage lexical diversity. Sampling uniformly from the vocabulary, however, would overrepresent rare words and underrepresent common ones, including function words. We therefore sample the word candidates using the frequency estimates from Stage~1, so that they better reflect natural patterns of word usage.

We split $\mathcal{V}$ at a rank boundary $b$ into high-frequency entries $\mathcal{H}$ and low-frequency entries $\mathcal{L}$ (Figure~\ref{fig:methodology}):
\begin{equation}
    \mathcal{H} = \{ w \in \mathcal{V} : r(w) \le b \}, \qquad
    \mathcal{L} = \mathcal{V} \setminus \mathcal{H}.
    \label{eq:hl_split}
\end{equation}
The word candidates should include entries from both sets. A composition parameter $c \in [0,1]$ controls the proportion of high-frequency entries: of the $k$ word candidates, $k_H = \lfloor c \cdot k \rfloor$ come from $\mathcal{H}$ and $k_L = k - k_H$ from $\mathcal{L}$.

Sampling low-frequency entries independently tends to yield unrelated words that are hard to combine into a coherent text. We therefore restrict them to a single topic so that they are semantically related. For each generation, we draw a topic $t$ uniformly from a fixed list of 20 topics.\footnote{The topics are food, family, holiday, travel, health, fitness, education, technology, music, movies, books, fashion, nature, pets, business, art, science, gaming, home, and [TOPIC].} We compute the semantic similarity $\mathrm{sim}(w, t)$ between the English equivalent $\mathrm{en}(w)$ of each entry $w \in \mathcal{L}$ and the topic $t$ using a sentence-transformer model,\footnote{\url{https://huggingface.co/sentence-transformers/all-MiniLM-L6-v2}} precomputing these scores once per topic. We then define $\mathcal{L}_t \subseteq \mathcal{L}$ as the $\lceil \rho \cdot |\mathcal{L}| \rceil$ entries with the highest similarity to $t$, where $\rho \in (0,1]$ is the topic fraction.

Within a subset $S \in \{\mathcal{H}, \mathcal{L}_t\}$, each entry is sampled with probability proportional to its estimated frequency:
\begin{equation}
    p(w \mid S) = \frac{f(w)}{\sum_{w' \in S} f(w')}, \qquad w \in S.
    \label{eq:sampling_prob}
\end{equation}
The candidate set $C$ is then
\begin{equation}
    C = C_{\mathcal{H}} \cup C_{\mathcal{L}_t}, \qquad |C| = k,
    \label{eq:candidates}
\end{equation}
where $C_{\mathcal{H}}$ consists of $k_H$ entries drawn from $\mathcal{H}$ and $C_{\mathcal{L}_t}$ of $k_L$ entries drawn from $\mathcal{L}_t$, both without replacement according to Equation~\eqref{eq:sampling_prob}, renormalized after each draw.

\paragraph{Stage 4: Assembling the Prompt.} 

Given $k$ and $s$, we first draw $s$ exemplars $e_1, \dots, e_s$ uniformly at random without replacement from $\mathcal{E}$.
Each exemplar $e_j$ is paired with its corresponding word candidate set $C_j=\mathcal{D}(e_j)$, where $\mathcal{D}(e_j)$ contains the dictionary entries provided when $e_j$ was created (Stage~2). Importantly, each exemplar uses only a subset of the provided candidates, demonstrating that the model should select words that fit naturally in context rather than force all candidates into the generated text. We then sample a new candidate set $C$ following Stage~3 and construct the prompt as \begin{equation}
    P = \big[\, (C_1, e_1);\ \dots;\ (C_s, e_s);\ C;\ I \,\big],
    \label{eq:prompt}
\end{equation}
where $[\,\cdot\,;\,\cdot\,]$ denotes concatenation and $I$ instructs the model to generate a new text using suitable words from $C$. We submit $P$ to an LLM with reasoning enabled, allowing the model to reason over the provided lexical and syntactic information before producing the final text, \[ y \sim \mathrm{LLM}(P). \] To generate $N$ synthetic texts, we repeat this procedure $N$ times, resampling the topic, few-shot exemplars, and candidate words for each generation.

\paragraph{Hyperparameters.} Our final configuration uses $k = 300$ word candidates, $s = 3$ exemplars, a rank boundary of $b = 500$, a composition of $c = 0.10$, and a topic fraction of $\rho = 0.25$. We selected these values through preliminary experiments in which we generated 5-word sentences and paragraphs under different settings (Appendix~\ref{sec:appendix_hyperparameter_settings}) and asked annotators to rate the fluency of the generated texts.

\section{Experimental Setup}
\label{sec:exp_setup}

\subsection{Configurations}
We conduct the experiment with 6 languages, 3 models, 2 thinking modes, 3 prompting strategies, and 5 text lengths, which results in a total of 540 combinations of configurations. We generate 100 outputs for each combination, yielding a total of 54,000 outputs. The configurations are as follows.

\paragraph{Languages.}
We experiment with six Southeast Asian languages spanning different levels of resource availability: Indonesian and Vietnamese as medium-resource languages, Javanese and Minangkabau as low-resource languages, and Acehnese and Hitu as extremely low-resource languages.

\paragraph{Models and Thinking Modes.}
We evaluate three LLMs: Gemini 3.5 Flash \citep{google2026gemini35flash}, Claude Haiku 4.5 \citep{anthropic2025haiku45}, and GPT-5.4 Mini \citep{openai2026gpt54mini}. Each model is evaluated with reasoning enabled and disabled, resulting in six configurations. We initially also considered open-weight models, including Llama 3.3 70B \citep{grattafiori2024llama3, meta2024llama33} and Qwen2.5 72B \citep{qwen2024qwen25}. However, preliminary evaluation showed substantially lower fluency, particularly for low-resource languages. Because our primary fluency evaluation requires native-speaker judgments, we exclude these models from the full evaluation and focus our annotation budget on the stronger model configurations.

\paragraph{Prompting Strategies.}
We compare three prompting strategies that differ in the amount of information provided to the LLM. The first two serve as baselines, while the third is the strategy used in our method:
\begin{enumerate}
\item \textbf{Topic-Conditioned Zero-Shot Prompt (Zero-shot):} This prompt only instructs the model to generate text in the target language about a given topic (see the full prompt in Appendix~\ref{sec:appendix_prompt_zero_shot}).
\item \textbf{Few-Shot Exemplars Only Prompt (Few-shot):} This prompt adds human-written exemplars of the target length to the instruction (see the full prompt in Appendix~\ref{sec:appendix_prompt_e_only}).
\item \textbf{Few-Shot Exemplars and Word Candidates Prompt (Ours):} This prompt adds both human-written exemplars of the target length and word candidates sampled from the dictionary to the instruction, as discussed in Section~\ref{sec:methodology} (see the full prompt in Appendix~\ref{sec:appendix_prompt_w+e}).
\end{enumerate}

\paragraph{Text Lengths.}
We generate texts at five sequence lengths: sentences of 5, 10, 15, and 20 words, and paragraphs consisting of 3 sentences and 25 words. This setup lets us test whether the LLM can generate text across a range of lengths.

\subsection{Evaluation Metrics.}
\label{subsec:eval_metrics}
We evaluate the generated outputs along three dimensions: fluency, off-target rate, and lexical diversity. We compute each score at the level of a configuration. 

\paragraph{Fluency.}
\label{subpar:human_fluency}
We conduct a human evaluation to assess the fluency of the generated texts. Due to annotation costs, we evaluate a subset of outputs covering three models with only thinking mode enabled, three prompting strategies, and three text lengths (5-word sentences, 15-word sentences, and paragraphs), with 30 texts per configuration, resulting in 810 texts per language. Each language is annotated by two annotators. They see only the generated texts, presented in a different random order for each annotator, and complete the annotation independently. They rate each text on a 1--5 scale following our annotation guidelines (see Appendix~\ref{sec:human_evaluation_guidelines}), where a higher score indicates a more fluent and natural text. We measure inter-annotator agreement using Pearson and Spearman correlations and observe moderate agreement between the two annotators (see Table~\ref{tab:correlations}). The final fluency score for each text is the average of the two annotators' ratings.

\begin{table}[t]
    \centering
    \begin{tabular}{l c c}
        \hline
        \textbf{Language} & \textbf{Pearson} & \textbf{Spearman} \\
        \hline
        Indonesian & 0.564 & 0.243 \\
        Vietnamese & 0.554 & 0.516 \\
        Javanese & 0.365 & 0.331 \\
        Minangkabau & 0.747 & 0.736 \\
        Acehnese & 0.774 & 0.771 \\
        Hitu & 0.466 & 0.494 \\
        \hline
    \end{tabular}
    \caption{Inter-annotator Pearson and Spearman correlations of human fluency ratings between the two annotators, by language.}
    \label{tab:correlations}
\end{table}

\paragraph{Off-Target Rate.}
We measure off-target rate as the proportion of words in a generated text that do not appear in the target-language dictionary. This metric evaluates a model’s tendency to produce text outside the desired language, reflecting language confusion~\citep{marchisio-etal-2024-understanding} or complete failure to produce that language. Scores range from 0 to 1, where 1 means that none of the generated text belongs to the target language, so higher values indicate greater language confusion. We score each output independently and average the scores over each configuration.

\paragraph{Lexical Diversity.}
We measure lexical diversity with a variant of distinct-$n$ that counts only $n$-grams composed of target-language words. Let $\mathcal{W}$ be the set of target-language words in the dictionary $\mathcal{D}$, and let $\mathcal{Y}$ be the set of texts generated under one configuration. Each text $y \in \mathcal{Y}$ is a sequence of words $(u_1, \dots, u_L)$, and an $n$-gram is a sequence of $n$ consecutive words $(u_i, \dots, u_{i+n-1})$ within a single text. We call an $n$-gram \textit{valid} if all of its words belong to $\mathcal{W}$. Let $G_n(\mathcal{Y})$ be the multiset of all $n$-grams in $\mathcal{Y}$, and let $U_n(\mathcal{Y}) \subseteq G_n(\mathcal{Y})$ be the set of distinct valid $n$-grams:
\begin{equation}
    U_n(\mathcal{Y}) = \{ g \in G_n(\mathcal{Y}) : g_1, \dots, g_n \in \mathcal{W} \},
    \label{eq:valid_ngrams}
\end{equation}
Lexical diversity is the ratio of distinct valid $n$-grams to all $n$-grams, averaged over $n$-gram lengths $1$ to $n_{\max}$:
\begin{equation}
    \mathrm{LD}(\mathcal{Y}) = \frac{1}{n_{\max}} \sum_{n=1}^{n_{\max}} \frac{|U_n(\mathcal{Y})|}{|G_n(\mathcal{Y})|},
    \label{eq:lexical_diversity}
\end{equation}
where we set $n_{\max} = 5$. The score lies in $[0, 1]$ and decreases with both repeated $n$-grams and off-target words, so higher values indicate a more diverse target-language vocabulary. We compute this score for each configuration.


\begin{table*}[t]
    \centering
    \small
    \setlength{\tabcolsep}{3pt}
    \renewcommand{\arraystretch}{1.1}
    \begin{tabular*}{0.92\textwidth}{@{\extracolsep{\fill}}ll ccc c ccccc@{}}
        \toprule
        \multirow{2}{*}{\textbf{\makecell{Language}}}
        & 
        \multirow{2}{*}{\textbf{\makecell{Prompt}}}
        & \multicolumn{3}{c}{\textbf{Fluency $\uparrow$}}
          & \multirow{2}{*}{\textbf{\makecell{Off-Target\\Rate $\downarrow$}}}
          & \multicolumn{5}{c}{\textbf{\begin{tabular}[c]{@{}c@{}}Lexical Diversity $\uparrow$\end{tabular}}} \\
        \cmidrule(lr){3-5} \cmidrule(l){7-11}
        & & 5 & 15 & Para. & & 5 & 10 & 15 & 20 & Para. \\
        \midrule
        \rowcolor{blue!20} \multicolumn{11}{c}{\textbf{Medium-Resource Languages}} \\
        Indonesian
          & Zero-shot & 4.989 & 4.950 & 4.761 & 0.028 & 0.542 & 0.573 & 0.570 & 0.588 & 0.546 \\
          & Few-shot  & \textbf{4.994} & \textbf{4.994} & \textbf{4.994} & 0.018 & 0.430 & 0.520 & 0.568 & 0.608 & 0.548 \\
          & Ours      & 4.972 & 4.911 & 4.811 & \textbf{0.014} & \textbf{0.868} & \textbf{0.859} & \textbf{0.834} & \textbf{0.817} & \textbf{0.721} \\
        \midrule
        Vietnamese
          & Zero-shot & 4.667 & 4.744 & \textbf{4.722} & 0.015 & 0.459 & 0.546 & 0.551 & 0.548 & 0.508 \\
          & Few-shot  & \textbf{4.856} & \textbf{4.828} & 4.522 & 0.004 & 0.457 & 0.512 & 0.523 & 0.545 & 0.448 \\
          & Ours      & 4.539 & 4.017 & 3.544 & \textbf{0.001} & \textbf{0.876} & \textbf{0.845} & \textbf{0.821} & \textbf{0.800} & \textbf{0.712} \\
        \rowcolor{yellow!20} \multicolumn{11}{c}{\textbf{Low-Resource Languages}} \\
        Javanese
          & Zero-shot & 3.956 & \textbf{4.239} & \textbf{4.394} & 0.252 & 0.266 & 0.282 & 0.263 & 0.273 & 0.242 \\
          & Few-shot  & \textbf{4.267} & 4.128 & 4.228 & 0.200 & 0.296 & 0.290 & 0.311 & 0.304 & 0.244 \\
          & Ours      & 3.744 & 3.728 & 3.811 & \textbf{0.090} & \textbf{0.733} & \textbf{0.632} & \textbf{0.609} & \textbf{0.604} & \textbf{0.470} \\
        \midrule
        Minangkabau
          & Zero-shot & 3.500 & 3.561 & 3.417 & 0.370 & 0.234 & 0.218 & 0.206 & 0.208 & 0.182 \\
          & Few-shot  & 3.894 & 3.789 & 3.578 & 0.221 & 0.344 & 0.334 & 0.315 & 0.293 & 0.256 \\
          & Ours      & \textbf{4.317} & \textbf{4.022} & \textbf{3.922} & \textbf{0.064} & \textbf{0.799} & \textbf{0.748} & \textbf{0.674} & \textbf{0.647} & \textbf{0.518} \\
        \rowcolor{red!20} \multicolumn{11}{c}{\textbf{Extremely Low-Resource Languages}} \\
        Acehnese
          & Zero-shot & 3.072 & 2.850 & 2.633 & 0.413 & 0.200 & 0.189 & 0.171 & 0.156 & 0.136 \\
          & Few-shot  & 3.639 & 3.239 & 2.739 & 0.315 & 0.234 & 0.230 & 0.213 & 0.198 & 0.192 \\
          & Ours      & \textbf{4.339} & \textbf{3.889} & \textbf{3.444} & \textbf{0.064} & \textbf{0.702} & \textbf{0.660} & \textbf{0.644} & \textbf{0.614} & \textbf{0.505} \\
        \midrule
        Hitu
          & Zero-shot & 1.789 & 1.661 & 1.611 & 0.836 & 0.025 & 0.017 & 0.016 & 0.015 & 0.013 \\
          & Few-shot  & 2.283 & 2.328 & 2.117 & 0.609 & 0.078 & 0.058 & 0.063 & 0.082 & 0.042 \\
          & Ours      & \textbf{3.394} & \textbf{2.867} & \textbf{2.828} & \textbf{0.060} & \textbf{0.747} & \textbf{0.670} & \textbf{0.639} & \textbf{0.615} & \textbf{0.509} \\
        \bottomrule
    \end{tabular*}
    \caption{Fluency, off-target rate, and lexical diversity per language and prompt. Each value is the average over models and thinking mode (for fluency, only models with thinking mode enabled). Best per language in \textbf{bold}.}
    \label{tab:main_results}
\end{table*}

\section{Results}
\label{sec:results}
Table~\ref{tab:main_results} reports the results for each language and prompt, averaged across all evaluated models. Results for each individual model and thinking mode are reported in Appendix~\ref{app:per_model_results} (Table~\ref{tab:per_model_results}). We organize the discussion around the evaluation metrics described in Section~\ref{subsec:eval_metrics}.


\paragraph{Fluency.}
Our prompt achieves the highest fluency for Minangkabau, Acehnese, and Hitu at every text length, exceeding the best baseline by an average of 0.33, 0.69, and 0.79 points, respectively. The margin grows as resource availability decreases and is largest for the extremely low-resource languages, where baseline fluency drops sharply (e.g., 1.61--1.79 for zero-shot in Hitu). For Indonesian, all prompts score close to the maximum, with few-shot marginally ahead. For Vietnamese and Javanese, however, our prompt trails both baselines, particularly for Vietnamese paragraphs (3.54 vs.\ 4.72 for zero-shot). We hypothesize that this stems from the quality of the underlying dictionaries: inaccurate or poorly matched entries are passed directly to the model, whereas the baselines rely only on its own knowledge. We examine this hypothesis in Section~\ref{sec:error-analysis-fluency}. Finally, fluency generally decreases as texts get longer.

\paragraph{Off-Target Rate.} 
Our prompt achieves the lowest off-target rate in all six languages. For both baselines, the rate rises steadily as resource availability decreases: with zero-shot prompting, it grows from 0.015--0.028 for the medium-resource languages to 0.836 for Hitu, where most generated words are not in the target language. Our prompt breaks this pattern, keeping the rate at or below 0.090 for every language. The reductions are therefore largest for the extremely low-resource languages, from 0.609 with few-shot prompting to 0.060 for Hitu and from 0.315 to 0.064 for Acehnese. For Indonesian and Vietnamese, all prompts produce very few off-target words, with differences of at most 0.014.

\paragraph{Lexical Diversity.}
Our prompt yields the highest lexical diversity for every language and text length. The gains hold even for the medium-resource languages, where off-target words are rare: for Indonesian, our prompt scores 0.72--0.87 against 0.43--0.61 for the baselines, indicating that the word candidates broaden the vocabulary rather than only reducing off-target words. The gap widens as resource availability decreases. For Hitu, the baselines score below 0.09 at every length because their outputs are largely off-target and repetitive, whereas our prompt scores 0.51--0.75, an average improvement of 0.57 over few-shot prompting. For our prompt, lexical diversity decreases steadily with text length in every language and is lowest for paragraphs, partly because longer texts inevitably repeat common words.

\begin{table}[t]
\centering
\begin{tabular}{lcccc}
\toprule
\textbf{Language} & \textbf{Spearman} & \textbf{MAE} & \textbf{Bias} \\
\midrule
Indonesian  & 0.37 & 0.12 & -0.04 \\
Vietnamese  & 0.39 & 0.44 & +0.40 \\
Javanese    & 0.34 & 0.87 & +0.44 \\
Minangkabau & 0.47 & 0.91 & +0.60 \\
Acehnese    & 0.69 & 0.86 & +0.58 \\
Hitu        & 0.15 & 1.70 & +1.19 \\
\bottomrule
\end{tabular}
\caption{Reliability of the LLM judge (Gemini 2.5 Flash) on fluency evaluation, measured against human ratings.}
\label{tab:judge-reliability}
\end{table}

\section{Analysis}
\label{sec:analysis}
\paragraph{Dictionary Qualities Hinder Fluency in Vietnamese and Javanese.}
\label{sec:error-analysis-fluency}
Our method scores below the baselines in fluency for Vietnamese and Javanese. To investigate these cases, we asked the annotators to examine the generated texts and the vocabulary entries provided to the model. Their feedback suggests that dictionary quality plays an important role. For Vietnamese, some dictionary entries were valid but uncommon in everyday usage, while more familiar alternatives were missing. As a result, semantically correct generations could still sound unnatural to native speakers. For Javanese, the dictionary primarily represents the Yogyakarta variety, whereas our annotator speaks Javanese with Arekan dialect. This mismatch led to unfamiliar vocabulary appearing in the generated text, reducing perceived fluency despite the words being valid dictionary entries. These observations highlight an important practical consideration for vocabulary-guided generation: the dictionary should be validated for both lexical coverage and alignment with the target language variety before large-scale generation. In particular, commonly used words should be adequately represented, and the lexical resource should match the variety spoken by the intended users.

\begin{table}[t]
    \centering
    \small
    \setlength{\tabcolsep}{6pt}
    \renewcommand{\arraystretch}{1.1}
    \begin{tabular}{ll cc}
        \toprule
        \textbf{Model} & \textbf{Thinking} & \textbf{\begin{tabular}[c]{@{}c@{}}Off-Target\\Rate $\downarrow$\end{tabular}} & \textbf{\begin{tabular}[c]{@{}c@{}}Lexical\\Diversity $\uparrow$\end{tabular}} \\
        \midrule
        GPT & Off & 0.347 & 0.311 \\
        & On & \textbf{0.322} & \textbf{0.341} \\
        \midrule
        Gemini & Off & 0.294 & 0.336 \\
        & On & \textbf{0.233} & \textbf{0.371} \\
        \midrule
        Claude & Off & \textbf{0.378} & \textbf{0.300} \\
        & On & 0.394 & 0.290 \\
        \bottomrule
    \end{tabular}
    \caption{Results per model and thinking mode, averaged over three languages (Hitu, Minangkabau, and Acehnese) and three prompts. Lexical diversity is averaged over the 5-, 10-, 15-, 20-word and paragraph settings. The best score between the thinking modes is shown in \textbf{bold}.}
    \label{tab:model_thinking_analysis}
\end{table}

\paragraph{Reliability of the LLM-as-a-Judge for Fluency Evaluation.}
We assess the reliability of the LLM-as-a-judge by comparing the fluency scores it assigns against the human fluency scores collected in our previous human evaluation (Section~\ref{subpar:human_fluency}) using the mean of the two annotators' scores across six languages. We use Gemini 2.5 Flash \citep{google2025gemini25flash} as the judge. For each language, let $h_i$ denote the mean of the human fluency scores and $j_i$ the judge's score for the $i$-th output, over $N$ outputs. We report Spearman's rank correlation $\rho$ between $h_i$ and $j_i$ to measure how well the judge preserves the human ranking of outputs, Mean Absolute Error (MAE) to measure how far its scores deviate from human scores, and mean signed bias to measure whether it over- or underestimates fluency.
\[
\begin{aligned}
\text{MAE} &= \frac{1}{N}\sum_{i=1}^{N}\left| j_i - h_i \right| \\
\text{Bias} &= \frac{1}{N}\sum_{i=1}^{N}\left( j_i - h_i \right)
\end{aligned}
\]

The judge's reliability varies across languages, as shown in Table~\ref{tab:judge-reliability}. Spearman's correlation is moderate for most languages, strongest for Acehnese (0.69), and weakest for Hitu (0.15), suggesting that the judge's rankings of Hitu texts align poorly with human judgments of fluency. MAE and bias show a clearer pattern.
For Indonesian, the judge's scores show the closest agreement with human ratings, with the lowest MAE (0.12) and a near-zero bias (-0.04). For the remaining languages, the MAE is generally higher, and the bias is consistently positive, indicating that the judge tends to assign higher scores than the human. This overestimation is particularly pronounced for Minangkabau (+0.60) and Acehnese (+0.58), and greatest for Hitu, which also has the highest MAE (+1.19).

Overall, the results show that the judge's agreement with human annotators varies across languages. While its fluency rankings align relatively well with human judgments for Acehnese, it tends to assign higher scores than human annotators and has larger scoring errors for several lower-resource languages. Therefore, LLM-as-a-judge should be used with caution for fluency evaluation in these languages and should not be considered a substitute for human evaluation.

\begin{figure}[t]
    \centering
    \begin{tikzpicture}
    \begin{axis}[
        width=0.95\columnwidth,
        height=6cm,
        xlabel={\# of generated sentences},
        ylabel={\% average unique valid n-gram},
        xmin=50, xmax=1050,
        ymin=0, ymax=120,
        xtick={100,200,300,400,500,600,700,800,900,1000},
        ytick={0,10,20,30,40,50,60,70,80,90,100},
        grid=major,
        legend style={
            at={(1,1)},
            anchor=north east,
            legend columns=1,
            font=\small
        },
        tick label style={font=\small},
        label style={font=\small},
        title style={font=\normalsize},
    ]

    \addplot[
        color=blue!60,
        dashed,
        mark=square*,
        mark size=2pt,
        line width=1pt
    ] coordinates {
        (100,82.946644)
        (200,80.180923)
        (300,76.844531)
        (400,74.793952)
        (500,73.062349)
        (600,71.938563)
        (700,70.565923)
        (800,69.302652)
        (900,68.366709)
        (1000,67.345222)
    };
    \addlegendentry{Ours}

    \addplot[
        color=orange!90!black,
        dashed,
        mark=square*,
        mark size=2pt,
        line width=1pt
    ] coordinates {
        (100,32.291763)
        (200,25.057402)
        (300,23.073825)
        (400,20.198296)
        (500,18.275706)
        (600,16.882821)
        (700,16.190034)
        (800,15.855958)
        (900,15.313696)
        (1000,14.597864)
    };
    \addlegendentry{Few-shot}

    \addplot[
        color=purple!60,
        dashed,
        mark=square*,
        mark size=2pt,
        line width=1pt
    ] coordinates {
        (100,36.143333)
        (200,28.188333)
        (300,25.472222)
        (400,21.985177)
        (500,20.185957)
        (600,19.302489)
        (700,17.949470)
        (800,16.804620)
        (900,15.960634)
        (1000,15.259422)
    };
    \addlegendentry{Zero-shot}

    \end{axis}
    \end{tikzpicture}
    \caption{Average percentage of unique valid n-grams as the number of generated Minangkabau sentences increases.}
    \label{fig:minangkabau-average-distinct-ngram}
\end{figure}

\paragraph{Thinking vs.\ Non-thinking Mode.}
Table~\ref{tab:model_thinking_analysis} compares the thinking mode between enabled and disabled mode for the three LRLs (Hitu, Minangkabau, and Acehnese). Enabling thinking tends to improve generation quality, but the effect depends on the model. Gemini 3.5 Flash benefits most, reducing the off-target rate by 0.06 and increasing lexical diversity by 0.03. GPT-5.4 Mini shows a smaller gain, with the off-target rate decreasing by 0.02 and lexical diversity increasing by 0.03. Claude Haiku 4.5 is the exception, as thinking brings no benefit. Overall, thinking helps in most cases, but the benefit is model-dependent.

\paragraph{Robustness Under Increasing Generation Volume.}
We investigate whether generating larger volumes of text leads to reduced lexical diversity and whether our prompt mitigates this effect compared with the baselines. We run this stress test experiment in Minangkabau and Hitu, generating up to 1,000 five-word sentences for each prompting strategy using Gemini 3.5 Flash with thinking mode enabled. We compute the percentage of unique valid n-grams among those observed in the first $N$ sentences, where $N = 100, 200, \ldots, 1000$. 

Figures~\ref{fig:minangkabau-average-distinct-ngram} and~\ref{fig:hitu-average-distinct-ngram} show that this percentage decreases as more sentences are generated across all prompting strategies, which is expected as repeated n-grams accumulate. In Minangkabau, the zero-shot and few-shot scores are 0.32 and 0.36 at 100 sentences and drop to between 0.14 and 0.15 at 1,000 sentences. The same pattern holds in Hitu, where the zero-shot and few-shot scores fall from 0.16 and 0.05 to 0.09 and 0.04, respectively. However, our prompt consistently maintains a higher percentage than the baselines in both languages, suggesting that it better preserves lexical diversity as generation volume increases.

\begin{figure}[t]
    \centering
    \begin{tikzpicture}
    \begin{axis}[
        width=0.95\columnwidth,
        height=6cm,
        xlabel={\# of generated sentences},
        ylabel={\% average unique valid n-gram},
        xmin=50, xmax=1050,
        ymin=0, ymax=120,
        xtick={100,200,300,400,500,600,700,800,900,1000},
        ytick={0,10,20,30,40,50,60,70,80,90,100},
        grid=major,
        legend style={
            at={(1,1)},
            anchor=north east,
            legend columns=1,
            font=\small
        },
        tick label style={font=\small},
        label style={font=\small},
        title style={font=\normalsize},
    ]

    \addplot[
        color=blue!60,
        dashed,
        mark=square*,
        mark size=2pt,
        line width=1pt
    ] coordinates {
        (100,71.856077)
        (200,66.989207)
        (300,62.545581)
        (400,59.784078)
        (500,58.359106)
        (600,56.053421)
        (700,54.680720)
        (800,52.897638)
        (900,51.650342)
        (1000,50.970602)
    };
    \addlegendentry{Ours}

    \addplot[
        color=orange!90!black,
        dashed,
        mark=square*,
        mark size=2pt,
        line width=1pt
    ] coordinates {
        (100,15.939613)
        (200,14.641699)
        (300,12.436261)
        (400,12.668663)
        (500,11.151295)
        (600,10.343158)
        (700,9.491967)
        (800,9.424020)
        (900,9.118445)
        (1000,8.915246)
    };
    \addlegendentry{Few-shot}

    \addplot[
        color=purple!60,
        dashed,
        mark=square*,
        mark size=2pt,
        line width=1pt
    ] coordinates {
        (100,4.912365)
        (200,4.120900)
        (300,3.852090)
        (400,3.583249)
        (500,3.374786)
        (600,3.489812)
        (700,3.552105)
        (800,3.341249)
        (900,3.468352)
        (1000,3.516287)
    };
    \addlegendentry{Zero-shot}

    \end{axis}
    \end{tikzpicture}
    \caption{Average percentage of unique valid n-grams as the number of generated Hitu sentences increases.}
    \label{fig:hitu-average-distinct-ngram}
\end{figure}

\section{Conclusion}

We introduced a training-free framework that leverages human-curated vocabularies, human-written exemplars, and LLM reasoning to generate synthetic text in low-resource languages. Across six languages, our framework produces more lexically diverse text with fewer off-target words than the baselines, while improving fluency particularly for the extremely low-resource Acehnese and Hitu. These results show the potential of human-curated vocabularies for bootstrapping synthetic text corpora where existing text resources are limited.

Our analysis also highlights important practical considerations. Dictionary quality and alignment with the target language variety can substantially affect generation quality, making vocabulary validation important before large-scale generation. The framework also requires additional input tokens for word candidates and exemplars, creating a trade-off between generation quality and cost. Future work should extend the framework to more low-resource languages and investigate whether the resulting synthetic corpora improve LLM pretraining and post-training.

\section*{Limitations}

Our experiments cover six Southeast Asian languages, five of which are Austronesian and all written in Latin script. Five are spoken in Indonesia and share considerable lexical material with Indonesian, a setting that is likely to favor positive transfer. We therefore cannot claim that the framework generalizes to non-Latin scripts or to agglutinative or polysynthetic morphology. 
We further describe Hitu as unseen during pretraining, but the training corpora of the three commercial models we use are not public, so this characterization is inferred from model behavior rather than established, as illustrated in Appendix~\ref{sec:appdx-output-sample}.

As our experiments are extensive, generating 54,000 texts made broad model coverage too costly, so we limit our study to three proprietary models accessed through commercial APIs, all from the economy tier of their respective families. Our findings are therefore restricted to these models, which are the latest versions of these model families at the time of our experiments. It remains unclear whether the gains hold for larger or more recent models, which may already encode more of these languages.

Furthermore, we evaluate the fluency, off-target rate, and lexical diversity of the generated text, but we have not trained or fine-tuned any model on the resulting data. With the quality gains in fluency and lexical diversity reported in Table~\ref{tab:main_results}, we see this as a promising direction for future work: testing whether our method benefits downstream models, particularly for LRLs.


\section*{Ethical Considerations}

This work involved native speakers of the six target languages, who wrote the few-shot exemplars (Appendix~\ref{sec:appendix_few_shot_exemplar_creation_procedure}) and rated the fluency of the generated texts (Appendix~\ref{sec:human_evaluation_guidelines}). Before agreeing to participate, annotators were informed of the purpose of the study, the expected workload, and the intended research use of their contributions. They were paid above the regional minimum wage in Indonesia and Vietnam and were free to withdraw at any point. No personal or sensitive information was collected, and none appears in our released materials. The Javanese, Minangkabau, Hitu, and Acehnese dictionaries are published by the Indonesian Ministry of Primary and Secondary Education, and the Indonesian and Vietnamese resources are distributed under Apache-2.0 and MIT licences; we use all resources for non-commercial academic research only and cite them in full (Appendix~\ref{sec:appendix_dictionary_resources}).

\section*{Acknowledgments}

This research was conducted through the apprenticeship program organized by the SEACrowd community. We are grateful to SEACrowd for facilitating this initiative and supporting emerging researchers in their early research journey by providing a platform for research collaboration and learning.


\bibliography{custom}

\appendix

\section{Dictionary Resources}
\label{sec:appendix_dictionary_resources}
The available resources vary across the languages used in our experiments. For Javanese\footnote{\href{https://kbji.kemendikdasmen.go.id/}{Digital Javanese-Indonesian Dictionary}} and Minangkabau\footnote{\href{https://limpapeh.kemendikdasmen.go.id/}{Minangkabau Language Dictionary}}, we collect dictionary entries through web scraping. Some entries in the Javanese dictionary lack information such as POS and example sentences. We use Gemini 3.5 Flash to fill in these missing fields based on the word and definition.
For Hitu\footnote{\href{https://repositori.kemendikdasmen.go.id/17075/}{Hitu–Indonesian Bilingual Dictionary}} and Acehnese\footnote{\href{https://repositori.kemendikdasmen.go.id/16244/}{Indonesian–Acehnese Bilingual Dictionary}}, the dictionary resources are available as PDF documents. We therefore use Gemini 3.5 Flash to extract the content from the scanned dictionary pages and convert it into a structured dataset.
For Indonesian\footnote{\href{https://huggingface.co/datasets/Lyon28/kamus-besar-bahasa-indonesia}{Lyon28/kamus-besar-bahasa-indonesia}} and Vietnamese\footnote{\href{https://huggingface.co/datasets/tsdocode/vietnamese-dictionary}{tsdocode/vietnamese-dictionary}}, we use an existing dictionary dataset available on Hugging Face. We further process the entries to obtain a consistent format across languages. For non-English definitions, we use Google Translate\footnote{\href{https://translate.google.com/}{Google Translate}} to translate them into English. We show the collected dictionary size in Table~\ref{tab:dictionary_size}.

\begin{table}[t]
    \centering
    \begin{tabular}{l c}
        \hline
        \textbf{Language} & \textbf{Number of Words} \\
        \hline
        Indonesian  & 71,017 \\
        Javanese    & 65,568 \\
        Vietnamese  & 33,968 \\
        Minangkabau & 33,149 \\
        Acehnese    & 10,638 \\
        Hitu        & 1,347 \\
        \hline
    \end{tabular}
    \caption{Number of words in the dictionary of each language.}
    \label{tab:dictionary_size}
\end{table}

\paragraph{Copyright and licensing.}
All resources are used solely for non-commercial academic research. The Javanese, Minangkabau, Hitu, and Acehnese dictionaries are published by the Indonesian Ministry of Primary and Secondary Education (Kemendikdasmen), and we cite the sources in full as permitted under Article~44 of the Indonesian Copyright Law (UU No.~28 Tahun 2014).\footnote{\href{https://peraturan.bpk.go.id/Read/28018/UU\%20Nomor\%2028\%20Tahun\%202014.pdf}{Law No. 28 of 2014 on Copyright}} The Indonesian and Vietnamese datasets are obtained from Hugging Face, where their licenses are stated as Apache-2.0 and MIT, respectively.

\section{Few-Shot Exemplar Creation Procedure}
\label{sec:appendix_few_shot_exemplar_creation_procedure}
We assigned one native-speaker annotator to each target language. Each annotator is a native speaker of or fluent in the target local language and has lived in the region where the language is spoken for a period of time. 

The exemplar creation process began with the development of annotation guidelines. Before commencing the task, annotators were asked to read the guidelines carefully and were encouraged to raise any questions or seek clarification about aspects they found unclear.

Each annotator was provided with a separate Google Sheets document to record the exemplars they created, along with the corresponding dictionary dataset to facilitate the selection of candidate words. The dictionary entries provided lexical information, including the word, part of speech, English definition, and example sentence. Annotators were instructed to use words from the provided dictionary dataset whenever possible when constructing the exemplars. A Jupyter Notebook was also provided to assist annotators in searching relevant dictionary entries. Figure~\ref{fig:guideline} presents the guidelines.

\begin{tcolorbox}[
    breakable,
    colback=gray!3,
    colframe=gray!50,
    boxrule=0.5pt,
    arc=2pt,
    left=8pt,
    right=8pt,
    top=8pt,
    bottom=8pt
]

\textbf{Exemplar Creation Guidelines}

\medskip

The annotation task consists of creating 40 sentences and 5 paragraphs in the
target language. The sentences should be distributed as follows:

\begin{itemize}
    \setlength{\itemsep}{2pt}
    \setlength{\parskip}{0pt}
    \item 10 sentences of approximately 5 words each;
    \item 10 sentences of approximately 10 words each;
    \item 10 sentences of approximately 15 words each;
    \item 10 sentences of approximately 20 words each; and
    \item 5 paragraphs, with each paragraph containing at least 3 sentences
          and at least 25 words.
\end{itemize}

Small variations in sentence length are acceptable as long as the sentence
remains natural. For example, a target of 5 words corresponds approximately
to 4--6 words, while a target of 10 words corresponds approximately to
9--11 words.

\medskip

For each exemplar, annotators should fill the following fields:

\begin{itemize}
    \setlength{\itemsep}{2pt}
    \setlength{\parskip}{0pt}
    \item \texttt{sentence}: the sentence or paragraph written in the target
          language;
    \item \texttt{translation}: the English translation of the sentence or
          paragraph; and
    \item \texttt{words\_candidates}: the dictionary entries corresponding
          to the words used to construct the sentence or paragraph.
\end{itemize}

For the \texttt{words\_candidates} field, annotators should identify the
dictionary entry corresponding to each word used in the sentence whenever
possible. Dictionary entries are provided in the following format:

\begin{center}
\texttt{word | pos | definition | example}
\end{center}

where \texttt{word} is the word in the target language, \texttt{pos} is its
part of speech, \texttt{definition} is its English definition, and
\texttt{example} is an example sentence when available.

\medskip

Annotators should use words from the provided dictionary dataset whenever possible. If an additional word is needed to produce a natural and
grammatically correct sentence but is not available in the dictionary
dataset, the annotator may introduce it as a new word candidate. The new
word should also be recorded using the same dictionary entry format.
\end{tcolorbox}

\captionof{figure}{Guidelines for creating few-shot exemplars.}
\label{fig:guideline}

\section{Hyperparameter Settings}
\label{sec:appendix_hyperparameter_settings}
We selected the final hyperparameters through preliminary experiments. For each candidate configuration, we generated 15 five-word sentences and paragraphs, and annotators rated the fluency of the generated texts. Table~\ref{tab:appendix_hyperparameter_settings} lists the hyperparameters and the values we searched over.
\begin{table}[H]
    \centering
    \small
    \setlength{\tabcolsep}{4pt}
    \label{tab:hyperparameter_settings}
    \begin{tabularx}{\columnwidth}{@{}c X l@{}}
        \toprule
        \textbf{Param.} & \textbf{Description} & \textbf{Values} \\
        \midrule
        $k$ & Number of word candidates   & $[100, 300, 500]$    \\
        $n$ & Number of few-shot examples & $[3, 5, 7, 10]$      \\
        $b$ & Rank boundary               & $[100, 500, 1000]$   \\
        $c$ & Composition                 & $[10\%, 20\%, 50\%]$ \\
        $p$ & Top-$p$ similarity fraction & $[25\%, 50\%]$       \\
        \bottomrule
    \end{tabularx}
    \caption{Hyperparameter settings.}
    \label{tab:appendix_hyperparameter_settings}
\end{table}

\section{Topic-Conditioned Zero-Shot Prompt}
\label{sec:appendix_prompt_zero_shot}
Figure~\ref{fig:topic_prompt} presents the prompt template used for topic-related generation. The \texttt{[TARGET\_LANGUAGE]}, \texttt{[OUTPUT\_TYPE]}, \texttt{[TOPIC]}, and \texttt{[TARGET\_SENTENCE\_LENGTH]} are filled by the target language (Hitu, Indonesian, Javanese, Minangkabau, Vietnamese, or Acehnese), the output type (\texttt{SENTENCE} or \texttt{PARAGRAPH}), a single topic, and the target sentence length (5, 10, 15, or 20 words), respectively. The topic is one of the following 20 topics: food, family, holiday, travel, health, fitness, education, technology, music, movies, books, fashion, nature, pets, business, art, science, gaming, home, and garden.

\begin{tcolorbox}[
    breakable,
    enhanced jigsaw,
    colback=gray!3,
    colframe=gray!50,
    boxrule=0.5pt,
    arc=2pt,
    left=8pt, right=8pt, top=8pt, bottom=8pt
]
\noindent
Generate a \texttt{[OUTPUT\_TYPE]} in the \texttt{[TARGET\_LANGUAGE]} language about \texttt{[TOPIC]}.

\medskip
\texttt{IF OUTPUT\_TYPE == SENTENCE:}\\
Ensure that the sentence is around \texttt{[TARGET\_SENTENCE\_LENGTH]} words long.\\
\texttt{IF OUTPUT\_TYPE == PARAGRAPH:}\\
Ensure that the paragraph contains at least 3 sentences and at least 25 words.

\medskip
Return only the \texttt{[OUTPUT\_TYPE]}.
\end{tcolorbox}
\captionof{figure}{Template for topic-conditioned zero-shot prompt.}
\label{fig:topic_prompt}

\section{Few-Shot Exemplars Only Prompt}
\label{sec:appendix_prompt_e_only}

Figure~\ref{fig:e_only_prompt} presents the prompt template used for generation with few-shot exemplars only, without word candidates. The \texttt{[TARGET\_LANGUAGE]}, \texttt{[OUTPUT\_TYPE]}, and \texttt{[TARGET\_SENTENCE\_LENGTH]} are filled by the target language, the output type (\texttt{SENTENCE} or \texttt{PARAGRAPH}), and the target sentence length (5, 10, 15, or 20 words), respectively. The few-shot examples (\texttt{\{FEW\_SHOT\_EXAMPLE\_1\}} to \texttt{\{FEW\_SHOT\_EXAMPLE\_N\}}) are $N$ sample outputs in the target language, which serve only as guidance for style, grammar, and fluency.

\begin{tcolorbox}[
    breakable,
    enhanced jigsaw,
    colback=gray!3,
    colframe=gray!50,
    boxrule=0.5pt,
    arc=2pt,
    left=8pt, right=8pt, top=8pt, bottom=8pt
]
\noindent
You are a linguistic expert in \texttt{[TARGET\_LANGUAGE]}.

\medskip
TASK: Create one coherent \texttt{[OUTPUT\_TYPE]} in \texttt{[TARGET\_LANGUAGE]}.

\medskip
===\\
RULES (MUST FOLLOW):\\
===

\medskip
1. FEW-SHOT GUIDANCE:\\
- Use the few-shot examples only as guidance for writing style, grammar, fluency, and natural sentence structure.\\
- Do not reuse the main entities, events, or semantic content from any few-shot example.\\
- Do not create output that is a paraphrase, extension, recombination, or slight modification of a few-shot example.

\medskip
2. PUNCTUATION \& CASING:\\
- You ARE allowed to use standard punctuation.\\
- You ARE allowed to capitalize the first letter of a word.

\medskip
3. LENGTH \& COVERAGE:\\
\texttt{IF OUTPUT\_TYPE == SENTENCE:}\\
- Output exactly one sentence.\\
- The sentence should be around \texttt{[TARGET\_SENTENCE\_LENGTH]} words.\\
\texttt{IF OUTPUT\_TYPE == PARAGRAPH:}\\
- Output a paragraph containing at least 3 sentences.\\
- The paragraph must contain at least 25 words.

\medskip
4. QUALITY:\\
- The output must be grammatically correct in \texttt{[TARGET\_LANGUAGE]}.\\
- Prioritize natural, coherent, and meaningful output.

\medskip
===\\
FEW-SHOT EXAMPLES (\textit{N} Examples)\\
===\\
Example 1: \texttt{\{FEW\_SHOT\_EXAMPLE\_1\}}\\
Example 2: \texttt{\{FEW\_SHOT\_EXAMPLE\_2\}}\\
\hspace*{8em}$\vdots$\\
Example \textit{N}: \texttt{\{FEW\_SHOT\_EXAMPLE\_N\}}\\
===\\
END OF EXAMPLES.\\
===

\medskip
Return only the \texttt{[OUTPUT\_TYPE]}.
\end{tcolorbox}
\captionof{figure}{Template for few-shot exemplars only prompt.}
\label{fig:e_only_prompt}

\section{Few-Shot Exemplars and Word Candidates Prompt}
\label{sec:appendix_prompt_w+e}
Figure~\ref{fig:w+e_prompt} presents the prompt template used for generation with few-shot exemplars and word candidates. The \texttt{[TARGET\_LANGUAGE]}, \texttt{[OUTPUT\_TYPE]}, and \texttt{[TARGET\_SENTENCE\_LENGTH]} are filled by the target language (Hitu, Indonesian, Javanese, Minangkabau, Vietnamese, or Acehnese), the output type (\texttt{SENTENCE} or \texttt{PARAGRAPH}), and the target sentence length (5, 10, 15, or 20 words), respectively. The few-shot examples consist of $N$ exemplars. Each exemplar $i$ contains a list of word candidates (\texttt{[WORD\_CANDIDATES\_$i$]}), formatted as \texttt{<word> | <part-of-speech> | <meaning> | <example\_usage>}, followed by its corresponding output in the target language (\texttt{[FEW\_SHOT\_EXAMPLE\_$i$]}). The final \texttt{[WORD\_CANDIDATES]} are the candidate words, in the same format, for which a new output is to be generated.

\begin{tcolorbox}[
    breakable,
    enhanced jigsaw,
    colback=gray!3,
    colframe=gray!50,
    boxrule=0.5pt,
    arc=2pt,
    left=8pt, right=8pt, top=8pt, bottom=8pt
]
\noindent
You are a linguistic expert in \texttt{[TARGET\_LANGUAGE]}.

\medskip
TASK: Create one coherent \texttt{[OUTPUT\_TYPE]} in \texttt{[TARGET\_LANGUAGE]} that sounds natural and fluent to a native speaker.

\medskip
===\\
RULES (MUST FOLLOW):\\
===

\medskip
1. VOCABULARY:\\
- Each entry in the ``Word Candidates'' list follows this structure, separated by `` | '':\\
<word> | <part-of-speech> | <meaning> | <example\_usage>\\
where:\\
- <word>: the \texttt{[TARGET\_LANGUAGE]} word to use.\\
- <part-of-speech>: the grammatical role of the word (e.g., noun, verb, adjective).\\
- <meaning>: the English meaning of the word.\\
- <example\_usage>: one or more example sentences in \texttt{[TARGET\_LANGUAGE]} showing how the word is naturally used, for reference only. If multiple examples are given, they are separated by `` * ''.\\
- Use the word candidates list as vocabulary guidance rather than a strict requirement.\\
- Try to use the candidate words where they fit naturally in the output.

\medskip
2. LENGTH \& COVERAGE:\\
\texttt{IF OUTPUT\_TYPE == SENTENCE:}\\
- Output exactly one sentence.\\
- The sentence should be around \texttt{[TARGET\_SENTENCE\_LENGTH]} words.\\
\texttt{IF OUTPUT\_TYPE == PARAGRAPH:}\\
- Output a paragraph containing at least 3 sentences.\\
- The paragraph must contain at least 25 words.

\medskip
3. PUNCTUATION \& CASING:\\
- You ARE allowed to use standard punctuation (periods, commas, question marks).\\
- You ARE allowed to capitalize the first letter of a word to start the output or for proper nouns.

\medskip
4. QUALITY:\\
- The output must be grammatically correct in \texttt{[TARGET\_LANGUAGE]}.\\
- DO NOT prioritize length over logical coherence. Prioritize natural, coherent, and meaningful output.\\
- DO NOT use specialized/technical terms unless they fit the context perfectly.

\medskip
5. ORIGINALITY:\\
- Do not reuse the main entities, events, or semantic content from any few-shot example.\\
- Do not create output that is a paraphrase, extension, recombination, or slight modification of a few-shot example.

\medskip
===\\
FEW-SHOT EXAMPLES (\textit{N} Examples)\\
===\\
Example 1\\
Word Candidates:\\
\texttt{[WORD\_CANDIDATES\_1]}\\
Output:\\
\texttt{[FEW\_SHOT\_EXAMPLE\_1]}

\medskip
Example 2\\
Word Candidates:\\
\texttt{[WORD\_CANDIDATES\_2]}\\
Output:\\
\texttt{[FEW\_SHOT\_EXAMPLE\_2]}

\medskip
$\vdots$

\medskip
Example \textit{N}\\
Word Candidates:\\
\texttt{[WORD\_CANDIDATES\_N]}\\
Output:\\
\texttt{[FEW\_SHOT\_EXAMPLE\_N]}\\
===\\
END OF EXAMPLES.\\
===

\medskip
Generate a new, unique \texttt{[OUTPUT\_TYPE]} based on the candidates below:\\
===\\
WORD CANDIDATES:\\
\texttt{[WORD\_CANDIDATES]}\\
===

\medskip
Return only the \texttt{[OUTPUT\_TYPE]}.
\end{tcolorbox}
\captionof{figure}{Template for few-shot exemplars and word candidates prompt.}
\label{fig:w+e_prompt}

\section{Human Evaluation Guidelines}
\label{sec:human_evaluation_guidelines}
Each annotator received a spreadsheet of the 810 generated texts for their language, with the texts in an \texttt{output} column, together with the instructions shown in Figure~\ref{fig:annotation_guidelines}.

\begin{tcolorbox}[
    breakable,
    enhanced jigsaw,
    colback=gray!3,
    colframe=gray!50,
    boxrule=0.5pt,
    arc=2pt,
    left=8pt, right=8pt, top=8pt, bottom=8pt
]
\noindent
\medskip
As a native speaker of \texttt{[TARGET\_LANGUAGE]}, your task is to rate the fluency of each given text on a 1--5 Likert scale.

\medskip
Scoring Rubric:\\
1. Very Poor: The text is largely unintelligible due to severe grammatical errors, broken syntax, or nonsensical word order. The intended meaning is difficult or impossible to determine. This score also applies if the text is an automated error response or is written in another language.\\
2. Poor: The intended meaning is mostly recoverable, but frequent grammatical errors, incorrect word order, or unnatural phrasing make the text difficult to read.\\
3. Fair: The meaning is clear, but the text contains some grammatical errors or awkward constructions that reduce its naturalness.\\
4. Good: The text is grammatically correct and easy to understand, with only minor stylistic awkwardness or slightly unnatural wording.\\
5. Excellent: The text is completely fluent, natural, and reads as if written by a proficient native speaker.

\medskip
Examples:\\
1. Very Poor: ``Yesterday market she because apples the quickly and was.''\\
2. Poor: ``She go market yesterday buy apples for she family.''\\
3. Fair: ``She went to market yesterday for buying apples for her family.''\\
4. Good: ``She went to the market yesterday to buy apples for her family.''\\
5. Excellent: ``She went to the market yesterday to buy fresh apples for her family, since they had run out at home.''

\end{tcolorbox}
\captionof{figure}{Instructions given to human annotators for the fluency evaluation.}
\label{fig:annotation_guidelines}

\section{Output Sample Comparison Between Baseline and Our Pipeline}
\label{sec:appdx-output-sample}
Table~\ref{tab:claude_illustration} illustrates how our method outperforms the baseline in Hitu using Claude Haiku 4.5. Without the few-shot exemplars and word candidates, the model fails to generate Hitu sentences and paragraphs because of its limited knowledge of the language. With the few-shot exemplars and the word candidates that offer vocabulary options (and without further training), the same model generates more fluent text in this language.

\begin{table}[h]
\centering
\small
\begin{tabularx}{\columnwidth}{l>{\raggedright\arraybackslash}Xc}
\toprule
\textbf{Language} & \textbf{Generated Text} & \textbf{Human} \\
& & \textbf{Score} \\
\midrule
\multicolumn{3}{l}{\textbf{5-word sentences}} \\
\midrule
Zero-shot  & \textit{I don't have reliable knowledge of the Hitu language to generate an accurate sentence. Hitu is a small Austronesian language spoken in Indonesia, and ...} & 1 \\
Ours  & Malole ma laka tarus nela. & 5 \\
\midrule
\multicolumn{3}{l}{\textbf{Paragraphs}} \\
\midrule
Zero-shot  & \textit{I appreciate your request, but I must be honest: I don't have reliable knowledge of the Hitu language to generate authentic text in it. Hitu is a language spoken in Indonesia's Maluku Islands ...} & 1 \\
Ours  & Baba laha ina una hetu ele luma asusu kula hatu pata maroto nale tolua. Sitewa manusia lalutu sataenai nai hahitin luma ma lama ingkuat hele. Ite manumpang kareta le lahan mahain ena supulumai tania kalu tawar heli opo palutan nyela. & 3 \\
\bottomrule
\end{tabularx}
\caption{The sample outputs of Claude Haiku 4.5 in zero-shot and our method at the 5-word sentence and paragraph level.}
\label{tab:claude_illustration}
\end{table}

\section{Information About AI Usage}
We used AI assistants only for polishing the writing and for literature search. We manually read and verified all papers found through AI-assisted search, and we take full responsibility for the content of this paper. All ideas originate from the authors.

\section{Detailed Results by Model and Thinking Mode}
\label{app:per_model_results}
Table~\ref{tab:per_model_results} reports fluency, off-target rate, and lexical diversity for each language, model, thinking mode, and prompt.

{\footnotesize
\setlength{\tabcolsep}{3pt}
\renewcommand{\arraystretch}{1.1}
\onecolumn
\begin{longtable}{llll ccc c ccccc}
\toprule
& & & & \multicolumn{3}{c}{\textbf{Fluency $\uparrow$}}
  & \multicolumn{1}{c}{\textbf{\begin{tabular}[c]{@{}c@{}}Off-Target Rate $\downarrow$\end{tabular}}}
  & \multicolumn{5}{c}{\textbf{\begin{tabular}[c]{@{}c@{}}Lexical Diversity $\uparrow$\end{tabular}}} \\
\cmidrule(lr){5-7} \cmidrule(lr){8-8} \cmidrule(l){9-13}
\textbf{Language} & \textbf{Model} & \textbf{Thinking} & \textbf{Prompt} & 5 & 15 & Para. & & 5 & 10 & 15 & 20 & Para. \\
\midrule
\endfirsthead
\multicolumn{13}{l}{\small\textit{\tablename~\thetable{} -- continued from previous page}} \\
\toprule
& & & & \multicolumn{3}{c}{\textbf{Fluency $\uparrow$}}
  & \multicolumn{1}{c}{\textbf{\begin{tabular}[c]{@{}c@{}}Off-Target Rate $\downarrow$\end{tabular}}}
  & \multicolumn{5}{c}{\textbf{\begin{tabular}[c]{@{}c@{}}Lexical Diversity $\uparrow$\end{tabular}}} \\
\cmidrule(lr){5-7} \cmidrule(lr){8-8} \cmidrule(l){9-13}
\textbf{Language} & \textbf{Model} & \textbf{Thinking} & \textbf{Prompt} & 5 & 15 & Para. & & 5 & 10 & 15 & 20 & Para. \\
\midrule
\endhead
\midrule
\multicolumn{13}{r}{\small\textit{Continued on next page}} \\
\endfoot
\endlastfoot
\rowcolor{blue!20} \multicolumn{13}{c}{\textbf{Medium-Resource Languages}} \\*
\midrule
Indonesian & GPT & On & Zero-shot & 4.967 & 4.933 & 4.400 & 0.036 & 0.577 & 0.541 & 0.578 & 0.606 & 0.565 \\*
 &  &  & Few-shot & \textbf{4.983} & \textbf{5.000} & \textbf{5.000} & \textbf{0.014} & 0.533 & 0.556 & 0.546 & 0.618 & 0.569 \\*
 &  &  & Ours & 4.967 & 4.917 & 4.700 & 0.015 & \textbf{0.888} & \textbf{0.857} & \textbf{0.824} & \textbf{0.817} & \textbf{0.715} \\
\cmidrule(l){3-13}
 &  & Off & Zero-shot & --- & --- & --- & 0.029 & 0.579 & 0.543 & 0.573 & 0.569 & 0.549 \\*
 &  &  & Few-shot & --- & --- & --- & \textbf{0.015} & 0.420 & 0.465 & 0.580 & 0.612 & 0.543 \\*
 &  &  & Ours & --- & --- & --- & \textbf{0.015} & \textbf{0.873} & \textbf{0.869} & \textbf{0.849} & \textbf{0.828} & \textbf{0.728} \\
\cmidrule(l){2-13}
 & Gemini & On & Zero-shot & \textbf{5.000} & \textbf{5.000} & 4.933 & 0.028 & 0.542 & 0.641 & 0.629 & 0.681 & 0.556 \\*
 &  &  & Few-shot & \textbf{5.000} & \textbf{5.000} & \textbf{5.000} & 0.025 & 0.420 & 0.656 & 0.673 & 0.650 & 0.546 \\*
 &  &  & Ours & 4.983 & 4.950 & 4.917 & \textbf{0.011} & \textbf{0.874} & \textbf{0.861} & \textbf{0.828} & \textbf{0.822} & \textbf{0.724} \\
\cmidrule(l){3-13}
 &  & Off & Zero-shot & --- & --- & --- & 0.029 & 0.454 & 0.473 & 0.429 & 0.491 & 0.511 \\*
 &  &  & Few-shot & --- & --- & --- & 0.016 & 0.191 & 0.240 & 0.261 & 0.395 & 0.530 \\*
 &  &  & Ours & --- & --- & --- & \textbf{0.013} & \textbf{0.869} & \textbf{0.869} & \textbf{0.847} & \textbf{0.826} & \textbf{0.730} \\
\cmidrule(l){2-13}
 & Claude & On & Zero-shot & \textbf{5.000} & 4.917 & 4.950 & 0.023 & 0.590 & 0.724 & 0.678 & 0.660 & 0.569 \\*
 &  &  & Few-shot & \textbf{5.000} & \textbf{4.983} & \textbf{4.983} & 0.023 & 0.600 & 0.686 & 0.719 & 0.697 & 0.619 \\*
 &  &  & Ours & 4.967 & 4.867 & 4.817 & \textbf{0.014} & \textbf{0.845} & \textbf{0.846} & \textbf{0.824} & \textbf{0.789} & \textbf{0.714} \\
\cmidrule(l){3-13}
 &  & Off & Zero-shot & --- & --- & --- & 0.025 & 0.510 & 0.517 & 0.536 & 0.519 & 0.526 \\*
 &  &  & Few-shot & --- & --- & --- & 0.018 & 0.416 & 0.516 & 0.631 & 0.676 & 0.481 \\*
 &  &  & Ours & --- & --- & --- & \textbf{0.016} & \textbf{0.858} & \textbf{0.850} & \textbf{0.832} & \textbf{0.817} & \textbf{0.716} \\
\midrule
Vietnamese & GPT & On & Zero-shot & 4.567 & 4.683 & \textbf{4.683} & 0.014 & 0.554 & 0.605 & 0.578 & 0.578 & 0.532 \\*
 &  &  & Few-shot & \textbf{4.850} & \textbf{4.817} & 4.417 & \textbf{0.001} & 0.447 & 0.507 & 0.453 & 0.516 & 0.452 \\*
 &  &  & Ours & 4.467 & 3.933 & 3.050 & \textbf{0.001} & \textbf{0.906} & \textbf{0.879} & \textbf{0.852} & \textbf{0.833} & \textbf{0.737} \\
\cmidrule(l){3-13}
 &  & Off & Zero-shot & --- & --- & --- & 0.014 & 0.427 & 0.514 & 0.535 & 0.550 & 0.535 \\*
 &  &  & Few-shot & --- & --- & --- & \textbf{0.001} & 0.377 & 0.384 & 0.322 & 0.435 & 0.374 \\*
 &  &  & Ours & --- & --- & --- & \textbf{0.001} & \textbf{0.900} & \textbf{0.855} & \textbf{0.857} & \textbf{0.830} & \textbf{0.743} \\
\cmidrule(l){2-13}
 & Gemini & On & Zero-shot & 4.900 & 4.900 & \textbf{4.933} & 0.017 & 0.399 & 0.639 & 0.667 & 0.613 & 0.493 \\*
 &  &  & Few-shot & \textbf{4.950} & \textbf{4.933} & 4.817 & 0.007 & 0.399 & 0.523 & 0.600 & 0.551 & 0.465 \\*
 &  &  & Ours & 4.817 & 4.417 & 4.133 & \textbf{0.002} & \textbf{0.861} & \textbf{0.830} & \textbf{0.797} & \textbf{0.767} & \textbf{0.683} \\
\cmidrule(l){3-13}
 &  & Off & Zero-shot & --- & --- & --- & 0.018 & 0.361 & 0.405 & 0.382 & 0.439 & 0.441 \\*
 &  &  & Few-shot & --- & --- & --- & 0.005 & 0.444 & 0.430 & 0.488 & 0.510 & 0.400 \\*
 &  &  & Ours & --- & --- & --- & \textbf{0.001} & \textbf{0.867} & \textbf{0.817} & \textbf{0.791} & \textbf{0.778} & \textbf{0.690} \\
\cmidrule(l){2-13}
 & Claude & On & Zero-shot & 4.533 & 4.650 & \textbf{4.550} & 0.015 & 0.558 & 0.644 & 0.654 & 0.621 & 0.538 \\*
 &  &  & Few-shot & \textbf{4.767} & \textbf{4.733} & 4.333 & 0.007 & 0.640 & 0.685 & 0.711 & 0.701 & 0.486 \\*
 &  &  & Ours & 4.333 & 3.700 & 3.450 & \textbf{0.002} & \textbf{0.841} & \textbf{0.828} & \textbf{0.802} & \textbf{0.790} & \textbf{0.711} \\
\cmidrule(l){3-13}
 &  & Off & Zero-shot & --- & --- & --- & 0.014 & 0.457 & 0.471 & 0.493 & 0.487 & 0.508 \\*
 &  &  & Few-shot & --- & --- & --- & 0.005 & 0.433 & 0.543 & 0.566 & 0.557 & 0.512 \\*
 &  &  & Ours & --- & --- & --- & \textbf{0.001} & \textbf{0.881} & \textbf{0.860} & \textbf{0.827} & \textbf{0.802} & \textbf{0.708} \\
\midrule
\rowcolor{yellow!20} \multicolumn{13}{c}{\textbf{Low-Resource Languages}} \\*
\midrule
Javanese & GPT & On & Zero-shot & 4.200 & \textbf{4.417} & \textbf{4.483} & 0.207 & 0.286 & 0.341 & 0.308 & 0.322 & 0.307 \\*
 &  &  & Few-shot & \textbf{4.550} & 4.217 & 4.317 & 0.153 & 0.300 & 0.354 & 0.358 & 0.353 & 0.288 \\*
 &  &  & Ours & 3.383 & 3.683 & 3.850 & \textbf{0.105} & \textbf{0.753} & \textbf{0.616} & \textbf{0.582} & \textbf{0.576} & \textbf{0.448} \\
\cmidrule(l){3-13}
 &  & Off & Zero-shot & --- & --- & --- & 0.217 & 0.314 & 0.295 & 0.295 & 0.300 & 0.288 \\*
 &  &  & Few-shot & --- & --- & --- & 0.141 & 0.367 & 0.308 & 0.316 & 0.343 & 0.304 \\*
 &  &  & Ours & --- & --- & --- & \textbf{0.090} & \textbf{0.736} & \textbf{0.642} & \textbf{0.604} & \textbf{0.577} & \textbf{0.486} \\
\cmidrule(l){2-13}
 & Gemini & On & Zero-shot & 4.550 & 4.150 & \textbf{4.583} & 0.259 & 0.196 & 0.277 & 0.265 & 0.286 & 0.228 \\*
 &  &  & Few-shot & \textbf{4.583} & \textbf{4.267} & 4.483 & 0.201 & 0.279 & 0.276 & 0.335 & 0.318 & 0.220 \\*
 &  &  & Ours & 3.983 & 3.850 & 3.900 & \textbf{0.096} & \textbf{0.744} & \textbf{0.604} & \textbf{0.635} & \textbf{0.594} & \textbf{0.427} \\
\cmidrule(l){3-13}
 &  & Off & Zero-shot & --- & --- & --- & 0.264 & 0.178 & 0.253 & 0.227 & 0.239 & 0.209 \\*
 &  &  & Few-shot & --- & --- & --- & 0.194 & 0.204 & 0.197 & 0.260 & 0.234 & 0.195 \\*
 &  &  & Ours & --- & --- & --- & \textbf{0.076} & \textbf{0.712} & \textbf{0.667} & \textbf{0.620} & \textbf{0.666} & \textbf{0.525} \\
 & Claude & On & Zero-shot & 3.117 & \textbf{4.150} & \textbf{4.117} & 0.279 & 0.322 & 0.294 & 0.247 & 0.262 & 0.220 \\*
 &  &  & Few-shot & 3.667 & 3.900 & 3.883 & 0.258 & 0.344 & 0.323 & 0.333 & 0.297 & 0.244 \\*
 &  &  & Ours & \textbf{3.867} & 3.650 & 3.683 & \textbf{0.089} & \textbf{0.751} & \textbf{0.624} & \textbf{0.586} & \textbf{0.599} & \textbf{0.436} \\
\cmidrule(l){3-13}
 &  & Off & Zero-shot & --- & --- & --- & 0.283 & 0.299 & 0.235 & 0.236 & 0.231 & 0.201 \\*
 &  &  & Few-shot & --- & --- & --- & 0.251 & 0.283 & 0.283 & 0.263 & 0.278 & 0.215 \\*
 &  &  & Ours & --- & --- & --- & \textbf{0.085} & \textbf{0.701} & \textbf{0.642} & \textbf{0.631} & \textbf{0.614} & \textbf{0.496} \\
\midrule
Minangkabau & GPT & On & Zero-shot & 3.967 & 3.733 & 3.467 & 0.310 & 0.291 & 0.240 & 0.239 & 0.236 & 0.250 \\*
 &  &  & Few-shot & 4.183 & 3.883 & 3.483 & 0.190 & 0.484 & 0.370 & 0.349 & 0.314 & 0.277 \\*
 &  &  & Ours & \textbf{4.367} & \textbf{4.150} & \textbf{3.833} & \textbf{0.059} & \textbf{0.820} & \textbf{0.763} & \textbf{0.701} & \textbf{0.647} & \textbf{0.520} \\
\cmidrule(l){3-13}
 &  & Off & Zero-shot & --- & --- & --- & 0.342 & 0.253 & 0.226 & 0.225 & 0.233 & 0.203 \\*
 &  &  & Few-shot & --- & --- & --- & 0.211 & 0.442 & 0.359 & 0.341 & 0.284 & 0.287 \\*
 &  &  & Ours & --- & --- & --- & \textbf{0.081} & \textbf{0.812} & \textbf{0.700} & \textbf{0.624} & \textbf{0.579} & \textbf{0.504} \\
\cmidrule(l){2-13}
 & Gemini & On & Zero-shot & 4.750 & 4.867 & 4.850 & 0.222 & 0.354 & 0.375 & 0.327 & 0.320 & 0.249 \\*
 &  &  & Few-shot & \textbf{4.933} & \textbf{4.900} & \textbf{4.950} & 0.122 & 0.347 & 0.447 & 0.412 & 0.383 & 0.320 \\*
 &  &  & Ours & 4.633 & 4.550 & 4.817 & \textbf{0.045} & \textbf{0.833} & \textbf{0.781} & \textbf{0.699} & \textbf{0.687} & \textbf{0.539} \\
\cmidrule(l){3-13}
 &  & Off & Zero-shot & --- & --- & --- & 0.252 & 0.262 & 0.276 & 0.283 & 0.296 & 0.244 \\*
 &  &  & Few-shot & --- & --- & --- & 0.149 & 0.266 & 0.340 & 0.365 & 0.353 & 0.330 \\*
 &  &  & Ours & --- & --- & --- & \textbf{0.062} & \textbf{0.731} & \textbf{0.776} & \textbf{0.675} & \textbf{0.665} & \textbf{0.553} \\
\cmidrule(l){2-13}
 & Claude & On & Zero-shot & 1.783 & 2.083 & 1.933 & 0.543 & 0.121 & 0.085 & 0.086 & 0.084 & 0.091 \\*
 &  &  & Few-shot & 2.567 & 2.583 & 2.300 & 0.336 & 0.294 & 0.246 & 0.233 & 0.206 & 0.166 \\*
 &  &  & Ours & \textbf{3.950} & \textbf{3.367} & \textbf{3.117} & \textbf{0.076} & \textbf{0.813} & \textbf{0.719} & \textbf{0.635} & \textbf{0.612} & \textbf{0.451} \\
\cmidrule(l){3-13}
 &  & Off & Zero-shot & --- & --- & --- & 0.552 & 0.122 & 0.106 & 0.074 & 0.082 & 0.058 \\*
 &  &  & Few-shot & --- & --- & --- & 0.318 & 0.234 & 0.238 & 0.188 & 0.221 & 0.158 \\*
 &  &  & Ours & --- & --- & --- & \textbf{0.061} & \textbf{0.787} & \textbf{0.747} & \textbf{0.713} & \textbf{0.693} & \textbf{0.540} \\
\midrule
\rowcolor{red!20} \multicolumn{13}{c}{\textbf{Extremely Low-Resource Languages}} \\*
\midrule
Acehnese & GPT & On & Zero-shot & 3.300 & 2.767 & 2.517 & 0.366 & 0.253 & 0.220 & 0.197 & 0.167 & 0.157 \\*
 &  &  & Few-shot & 3.817 & 3.083 & 2.417 & 0.307 & 0.338 & 0.262 & 0.212 & 0.183 & 0.158 \\*
 &  &  & Ours & \textbf{4.367} & \textbf{3.750} & \textbf{3.133} & \textbf{0.058} & \textbf{0.810} & \textbf{0.736} & \textbf{0.700} & \textbf{0.639} & \textbf{0.454} \\
\cmidrule(l){3-13}
 &  & Off & Zero-shot & --- & --- & --- & 0.377 & 0.242 & 0.233 & 0.194 & 0.177 & 0.176 \\*
 &  &  & Few-shot & --- & --- & --- & 0.303 & 0.276 & 0.293 & 0.227 & 0.216 & 0.202 \\*
 &  &  & Ours & --- & --- & --- & \textbf{0.095} & \textbf{0.760} & \textbf{0.638} & \textbf{0.603} & \textbf{0.562} & \textbf{0.436} \\
\cmidrule(l){2-13}
 & Gemini & On & Zero-shot & 4.600 & \textbf{4.650} & 4.017 & 0.264 & 0.273 & 0.295 & 0.265 & 0.252 & 0.210 \\*
 &  &  & Few-shot & \textbf{4.817} & 4.483 & 4.283 & 0.206 & 0.256 & 0.288 & 0.324 & 0.269 & 0.263 \\*
 &  &  & Ours & 4.733 & 4.583 & \textbf{4.483} & \textbf{0.072} & \textbf{0.537} & \textbf{0.587} & \textbf{0.576} & \textbf{0.564} & \textbf{0.512} \\
\cmidrule(l){3-13}
 &  & Off & Zero-shot & --- & --- & --- & 0.365 & 0.162 & 0.190 & 0.210 & 0.177 & 0.177 \\*
 &  &  & Few-shot & --- & --- & --- & 0.276 & 0.146 & 0.250 & 0.239 & 0.234 & 0.237 \\*
 &  &  & Ours & --- & --- & --- & \textbf{0.072} & \textbf{0.570} & \textbf{0.620} & \textbf{0.593} & \textbf{0.601} & \textbf{0.531} \\
\cmidrule(l){2-13}
 & Claude & On & Zero-shot & 1.317 & 1.133 & 1.367 & 0.561 & 0.126 & 0.100 & 0.076 & 0.084 & 0.055 \\*
 &  &  & Few-shot & 2.283 & 2.150 & 1.517 & 0.395 & 0.198 & 0.145 & 0.136 & 0.158 & 0.151 \\*
 &  &  & Ours & \textbf{3.917} & \textbf{3.333} & \textbf{2.717} & \textbf{0.052} & \textbf{0.765} & \textbf{0.659} & \textbf{0.665} & \textbf{0.611} & \textbf{0.513} \\
\cmidrule(l){3-13}
 &  & Off & Zero-shot & --- & --- & --- & 0.546 & 0.144 & 0.098 & 0.086 & 0.077 & 0.042 \\*
 &  &  & Few-shot & --- & --- & --- & 0.403 & 0.193 & 0.143 & 0.142 & 0.126 & 0.139 \\*
 &  &  & Ours & --- & --- & --- & \textbf{0.036} & \textbf{0.770} & \textbf{0.720} & \textbf{0.724} & \textbf{0.710} & \textbf{0.586} \\
\midrule
Hitu & GPT & On & Zero-shot & 1.783 & 1.550 & 1.517 & 0.890 & 0.015 & 0.011 & 0.007 & 0.009 & 0.005 \\*
 &  &  & Few-shot & 2.217 & 2.167 & 1.800 & 0.672 & 0.054 & 0.018 & 0.033 & 0.054 & 0.013 \\*
 &  &  & Ours & \textbf{3.200} & \textbf{2.917} & \textbf{2.750} & \textbf{0.049} & \textbf{0.775} & \textbf{0.699} & \textbf{0.626} & \textbf{0.590} & \textbf{0.459} \\
\cmidrule(l){3-13}
 &  & Off & Zero-shot & --- & --- & --- & 0.892 & 0.014 & 0.011 & 0.009 & 0.007 & 0.007 \\*
 &  &  & Few-shot & --- & --- & --- & 0.676 & 0.074 & 0.046 & 0.027 & 0.063 & 0.016 \\*
 &  &  & Ours & --- & --- & --- & \textbf{0.144} & \textbf{0.645} & \textbf{0.521} & \textbf{0.464} & \textbf{0.472} & \textbf{0.302} \\
 & Gemini & On & Zero-shot & 2.583 & 2.433 & 2.317 & 0.695 & 0.077 & 0.049 & 0.049 & 0.043 & 0.041 \\*
 &  &  & Few-shot & 2.567 & 2.683 & 2.783 & 0.433 & 0.149 & 0.126 & 0.120 & 0.127 & 0.099 \\*
 &  &  & Ours & \textbf{3.583} & \textbf{3.050} & \textbf{2.950} & \textbf{0.039} & \textbf{0.752} & \textbf{0.679} & \textbf{0.664} & \textbf{0.614} & \textbf{0.556} \\
\cmidrule(l){3-13}
 &  & Off & Zero-shot & --- & --- & --- & 0.777 & 0.043 & 0.031 & 0.028 & 0.031 & 0.026 \\*
 &  &  & Few-shot & --- & --- & --- & 0.642 & 0.078 & 0.061 & 0.066 & 0.077 & 0.074 \\*
 &  &  & Ours & --- & --- & --- & \textbf{0.050} & \textbf{0.739} & \textbf{0.671} & \textbf{0.656} & \textbf{0.629} & \textbf{0.552} \\
\cmidrule(l){2-13}
 & Claude & On & Zero-shot & 1.000 & 1.000 & 1.000 & 0.881 & 0.000 & 0.000 & 0.000 & 0.000 & 0.000 \\*
 &  &  & Few-shot & 2.067 & 2.133 & 1.767 & 0.663 & 0.044 & 0.036 & 0.051 & 0.079 & 0.025 \\*
 &  &  & Ours & \textbf{3.400} & \textbf{2.633} & \textbf{2.783} & \textbf{0.036} & \textbf{0.779} & \textbf{0.719} & \textbf{0.733} & \textbf{0.680} & \textbf{0.615} \\
\cmidrule(l){3-13}
 &  & Off & Zero-shot & --- & --- & --- & 0.879 & 0.000 & 0.000 & 0.000 & 0.000 & 0.000 \\*
 &  &  & Few-shot & --- & --- & --- & 0.566 & 0.069 & 0.065 & 0.083 & 0.095 & 0.027 \\*
 &  &  & Ours & --- & --- & --- & \textbf{0.041} & \textbf{0.790} & \textbf{0.734} & \textbf{0.688} & \textbf{0.707} & \textbf{0.572} \\
\bottomrule
\caption{Fluency, off-target rate, and lexical diversity per language, model, thinking mode, and prompt. Fluency was only human-evaluated for models with thinking mode enabled (--- = not evaluated). Best per model and thinking mode in \textbf{bold}.}
\label{tab:per_model_results}
\end{longtable}
}

\end{document}